\documentclass[11pt,a4paper]{article}
\usepackage[hyperref]{acl2019}
\usepackage{hyperref}
\usepackage{times}
\usepackage{latexsym}
\usepackage{url}
\usepackage[justification=centering]{caption}
\usepackage{graphicx}
\usepackage{tabularx}
\usepackage{soul}
\usepackage{algorithm,algorithmic}
\usepackage{textgreek}
\usepackage{multirow}
\usepackage{url}
\usepackage{enumitem}
\usepackage{float}
\usepackage{tabu}
\usepackage{array}
\usepackage{booktabs}
\usepackage{mathtools}
\usepackage{tikz}
\usepackage[normalem]{ulem}
\usepackage{gensymb}
\usepackage{amssymb}
\usepackage{pifont}
\usepackage{tikz}
\usepackage[export]{adjustbox}
\setlist{noitemsep}

\newcommand\blfootnote[1]{%
  \begingroup
  \renewcommand\thefootnote{}\footnote{#1}%
  \addtocounter{footnote}{-1}%
  \endgroup
}

\newcommand{\footlabel}[2]{%
    \addtocounter{footnote}{1}%
    \footnotetext[\thefootnote]{%
        \addtocounter{footnote}{-1}%
        \refstepcounter{footnote}\label{#1}%
        #2%
    }%
    $^{\ref{#1}}$%
}

\newcommand{\footref}[1]{%
    $^{\ref{#1}}$%
}

\aclfinalcopy % Uncomment this line for the final submission
\title{Code-Mixed to Monolingual Translation Framework}

\author{
        Sainik Kumar Mahata\textsuperscript{1}, Soumil Mandal\textsuperscript{2}, Dipankar Das\textsuperscript{3}, Sivaji Bandyopadhyay\textsuperscript{4} \\ \\
        \textsuperscript{1,3,4}Jadavpur University, Kolkata, India\\
       	\textsuperscript{2}SRM University, Chennai, India\\
        \textcolor{black!50}{sainik.mahata@gmail.com, soumil.mandal@gmail.com}\\
        \textcolor{black!50}{dipankar.dipnil2005@gmail.com, sivaji\_cse\_ju@yahoo.com}                
}

\date{}

\begin{document}
\maketitle
\begin{abstract}
The use of multi-lingualism in the new generation is widespread in the form of code-mixed data on social media, and therefore a robust translation system is required for catering to the monolingual users, as well as for easier comprehension by language processing models. In this work, we present a translation framework that uses translation-transliteration strategy for translating code-mixed data into their equivalent monolingual instances. For converting the output to a more fluent form, it is reordered using a target language model. The most important advantage of the proposed framework is that it does not require a code-mixed to monolingual parallel corpus at any point. On testing the framework, it achieved BLEU and TER scores of 16.47 and 55.45, respectively. Since the proposed framework deals with various sub-modules, we dive deeper into the importance of each of them, analyze the errors and finally, discuss some improvement strategies. \blfootnote{1, 2 equal contribution}
\end{abstract}

\section{Introduction}
\label{intro}
India has a linguistically diverse diaspora due to its long history of foreign acquaintances. English, one of those borrowed languages, became an integral part of the education system and hence gave rise to a population who are very comfortable using bilingualism in communication. This kind of language diversity and dialects initiates frequent code-mixing. Further, due to the emergence of social media, the practice has become even more widespread. We found out that only 26\%\footnote{\url{https://en.wikipedia.org/wiki/Multilingualism_in_India}} of the Indian population are bilingual. To cater to the rest, who are comfortable using only one native language and to make them compatible in the age of social media, translating code-mixed data into its corresponding monolingual instance is an alternative. But, translating such data manually requires a lot of effort and hence availing machines for the same is more desirable.

Machine translation it self is a challenging task due to out of vocabulary problems, context misunderstanding, grammatical errors, bias, etc. Thus, it becomes more difficult when the input instance is code-mixed, as many new challenges emerge with it. In this work, we present an architecture for code-mixed translation which doesn't require a code-mixed to monolingual parallel corpus for training. This is highly beneficial as code-mixed data is difficult to scrape and an enormous amount of data would be required for a model like SMT or NMT to learn the nuances of the language in order to properly translate. We implemented our architecture for Bengali-English (Bn-En) code-mixed data in Roman script to Bengali in Devanagari script. Our architecture is capable of translating monolingual sentences as well, for example in our case, if the input is in monolingual Bengali or English in Roman script, it will still translate it to the target language, which is Bengali in Devanagari. Our contributions also include preparation of a gold standard Bn-En code-mixed to Bn parallel corpus which was used for testing purposes only. The shortcomings and errors have been analyzed in detail as well.

\section{Related Work}
Several research works has been done in the recent past on code-mixed data, and especially involving language tagging. \citet{jhamtani:2014word} created an ensemble model by combining two classifiers to create a Hindi-English code-mixed LID. The first classifier used word frequency, modified edit distance, and character n-grams as features. The second classifier used the output from the former classifier for the current word, along with language and POS tag of neighbouring words to give the final tag. \citet{rijhwani:2017estimating} proposed a generalized language tagger for arbitrary set of languages which is fully unsupervised. With respect to back-transliteration,  \citet{bilac:2004hybrid} proposed a hyrbid approach which combines phoneme, grapheme and segmentation based modules. \citet{luo:2015handling} presented an architecture for back transliteration using an SMT framework described in \cite{Franz03:statisticalphrase-based}. \citet{ravishankar:2017finite} describes a finite-state based system for back-transliteration of transliterated Marathi words in Roman. The major advantage over statistical models is that its able to model exceptions without being retrained. \citet{sinha:2005machine} took the challenge of translation of Hindi-English code-mixed to English monolingual from a linguistics point of view by using morphological analyzers though they did not perform any in depth analysis or evaluations. In \cite{dhar:2018enabling}, the authors created a code-mixed (Hindi-English) to monolingual (English) parallel corpus consisting of 6096 instances. They also developed an augmentation pipeline which can be utilized for augmenting existing MT systems such that the translation of the systems can be improved without training the MT system specifically for code-mixed text. On testing the module with Moses, Google NMTS and Bing translator, the BLEU scores improved by 2\%, 9.4\% and 6.1\% respectively. To the best of our knowledge, ours is the first end-to-end code-mixed translation system.

\section{Parallel Corpus}
In order to build our test data, we randomly collected 1600 code-mixed instance from the En-Bn data prepared in \cite{patra:2018sentiment}. For creating the parallel corpus, a group consisting of three annotators who were fluent in both English and Bengali were employed. One of the annotators was asked to translate all the instances, while the other two classified the translations into two classes, correct and incorrect. The agreement was then calculated using Fleiss' Kappa \cite{fleiss:1973equivalence}, which was found to be $\approx$ 0.85. It is to be noted that the language tagger and back-transliteration models were created from resources different from our testing data.  

\section{Architecture}
\begin{figure}[ht]
\centering
\includegraphics[scale=0.34]{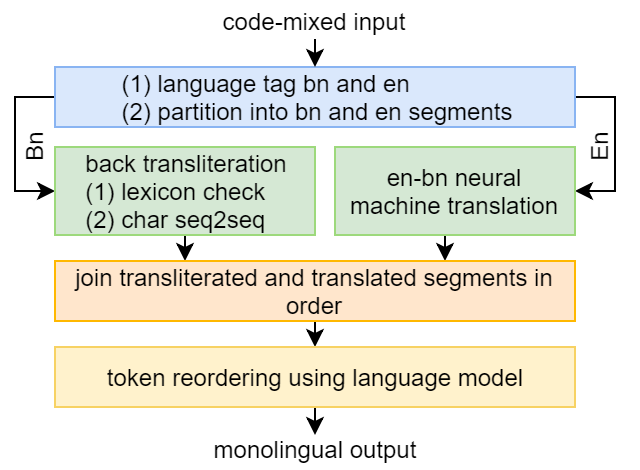}
\caption{Architecture overview.}
\label{fig1}
\end{figure}

Our proposed approach comprises of four modules. The first module is the language identification system that helps us to segment (boundaries of sub-sequences that are in same language) a code mixed sentence. The second module translates the English segments to Bengali using a character based neural machine translation system. Bengali segments written in Roman are back-transliterated to Devanagari form by the third module. Joining the translated and the back-transliterated segments into a monolingual instance, we noticed that the output wasn't always fluent, and had grammatical errors. To counter this, we developed the fourth module, that uses a language modelling to convert the output to a more natural looking instance with better flow. The architecture is depicted in Figure \ref{fig1}. All the models are described in detail below.

\subsection{Language Tagging \& Segmentation}
\label{segment}
This module partitions the input into segments with respect to language. Bn tagged segments are passed to the transliteration system while En segments are passed to the translation system. In our case, segments are sub-sequences of the instance, written in the same language. Strings in brackets denote segments.
\begin{flushleft}
\textit{E.g 1.} (Movie)\textsubscript{En} (ta bhalo chilo)\textsubscript{Bn} (but mid point)\textsubscript{En} (e amar khub)\textsubscript{Bn} (boring)\textsubscript{En} (lagte shuru korlo)\textsubscript{Bn}. \\
\textit{E.g 2.} (I had to go)\textsubscript{En} (karon o khub)\textsubscript{Bn} (urgently)\textsubscript{En} (daklo amaye)\textsubscript{Bn}.
\end{flushleft}
\noindent In order to achieve this goal, a language tagger was used. We used the character based LSTM architecture proposed by \citet{mandal:2018language}. This is a model having stacked LSTM of sizes 15-35-25-1, in order where 15 is the input dimension while 1 is the output dimension. The data used for training and testing were gathered from the data released in ICON 16~\footnote{http://ltrc.iiit.ac.in/icon2016/} and \citet{mandal2018analyzing}. Training dataset contains 6,632 words of Bn and En type each while test dataset comprises 700 words of Bn and En type each. With respect to our present experiment, the training data was increased by 1,400 sentences for both English and Bengali. This was collected from the code-mixed data released in \citet{ghosh2017sentiment}. Sources of all these instances, as described in their papers, were from social media websites like Twitter, Facebook and WhatsApp. The Loss function used was binary cross-entropy and we employ adam optimizer with sigmoid activation function. Epochs was set to 500 and batch size at 256. The increase in size of training data resulted in improvement in accuracy from 91.71\%, as was shown \citet{mandal:2018language}, to 93.2\%, when experimented on identical test data. 

\subsection{Back Transliteration}
\label{transliterate}
To make an accurate back-transliteration system, we used two resources, namely BN\_TRANS and PL which is described in \citet{mandal:2018normalization}. BN\_TRANS is essentially a parallel lexicon with two columns where col\_1 has Bn words in native script while col\_2 has the respective ITRANS~\footnote{https://en.wikipedia.org/wiki/ITRANS} transliterations. PL is a parallel lexicon where col\_1 has phonetically transliterated Bn words in Roman, while col\_2 has the respective ITRANS form. BN\_TRANS has 21850 entries in each column while PL has 6000 entries in each column.\\

\begin{figure}[ht]
\centering
\includegraphics[scale=0.34]{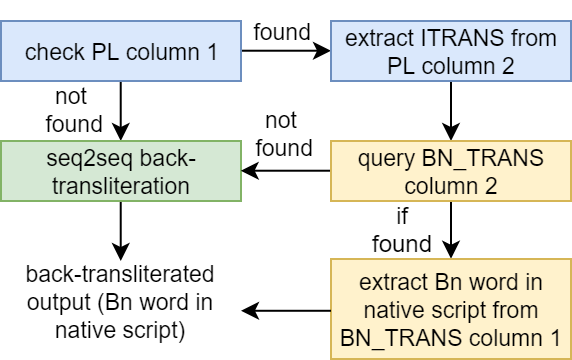}
\caption{Back-transliteration algorithm.}
\label{fig2}
\end{figure}

\noindent Our back transliteration system first performs lexical checking, i.e. it checks if the word is present in PL\textsubscript{col\_1}. If yes, it takes the respective ITRANS form and queries BN\_TRANS, i.e. it checks if it is present in BN\_TRANS\textsubscript{col\_2} and returns the respective word in native script. As there are several possible cases where the words are absent in PL, i.e. out of vocabulary, we decided to make a back transliteration system using character based seq2seq model \cite{ling:2015character} in order to resolve this scenario. We simply used BN\_TRANS as a parallel lexicon, where column\_2 entries are source sequences while column\_1 entries are target sequences, i.e. the goal of our model is to essentially learn the mappings from Bn in Roman script to Bn in native script. For training, the activation function used was softmax, optimizer was rmsprop, and loss function was categorical cross-entropy. Size of latent dimensions was set at 128, batch size was kept at 64, and number of epochs was set to 100. The training accuracy at the end was 48.2\%. The architecture is shown in Fig~\ref{fig2}

\subsection{English-Bengali Translation}
\label{translation}
For translating the English segments to its corresponding Bengali script, we decided to go for fully character level neural machine translation based on the architecture described in \citet{lee:2017fully} as it outperforms a statistical model \cite{mahata:2018smt}. It relies on the sequence-to-sequence \cite{sutskever:2014sequence} model and uses attention mechanism \cite{vaswani:2017attention} while decoding. We opted for this because of the benefits it provides over word level which are very important in our case. The benefits as stated in \citet{chung:2016character} are (1) capability to model morphological variants (2) overcomes out-of-vocabulary issue (3) do not require segmentation. \\

\noindent The seq2seq model takes a sequence \textit{X} = \{x\textsubscript{1}, x\textsubscript{2}, ..., x\textsubscript{n}\} as input and tries to generate the target sequence \textit{Y} = \{y\textsubscript{1}, y\textsubscript{2}, ..., y\textsubscript{m}\} as output, where x\textsubscript{i} and y\textsubscript{i} are the input and target symbols respectively. The architecture of seq2seq model comprises of two parts, the encoder and decoder. In order to build the encoder, we used LSTM cells. The input of the cell was one hot tensor of English sentences (embedding at character level). From the encoder, the internal states of each cell were preserved and the outputs were discarded. The purpose of this is to preserve the information at context level. These states were then passed on to the decoder cell as initial states. For building the decoder, again an LSTM cell was used with initial states as the hidden states from encoder. It was designed to return both sequences and states. The input to the decoder was one hot tensor (embedding at character level) of Bengali and Hindi sentences while the target data was identical, but with an offset of one time-step ahead.  The information for generation is gathered from the initial states passed on by the encoder. Thus, the decoder learns to generate target data [t+1,...] given targets [..., t] conditioned on the input sequence. It essentially predicts the output sequence, one character per time step. \\

\noindent For training and testing, the En-Bn parallel from TDIL~\footlabel{TDIL}{http://tdil.meity.gov.in/} and the corpus in \citet{post:2012constructing} was divided into 180k and 20k instances respectively. For training the model, batch size was set to 64, number of epochs was set to 100, activation function was softmax, optimizer chosen was rmsprop and loss function used was categorical cross-entropy. Learning rate was set to 0.001. Post training, the BLEU score of the model was calculated to be 5.06. 

% \begin{table*}[ht]
% \centering
% \begin{tabular}{|c|c|c|}
% \hline
% \textbf{code-mixed input} & \textbf{o/p without token reordering} & \multicolumn{1}{l|}{\textbf{o/p with token reordering}} \\ \hline
% \begin{tabular}[c]{@{}c@{}} \small{Janina how that is possible}\\ \small{(\textit{Don't know how that is possible})}\end{tabular} & \myfont{জানিনা কিভাবে \colorbox{pink}{সম্ভব যে}}  & \myfont{জানিনা কিভাবে \colorbox{lime}{যে সম্ভব}} \\ \hline
% \begin{tabular}[c]{@{}c@{}} \small{Ora repeatedly bollo don’t go}\\ \small{(\textit{They repeatedly told us not to go})} \end{tabular} & \myfont{ওরা বারবার বললো \colorbox{pink}{হবে না যেতে}} & \myfont{ওরা বারবার বললো \colorbox{lime}{যেতে হবে না}} \\ \hline
% \end{tabular}
% \caption{Translation with and without token reordering on short snippets.}
% \label{sample_io}
% \end{table*}

\begin{figure*}[!htb]
\centering
\includegraphics[scale=0.95]{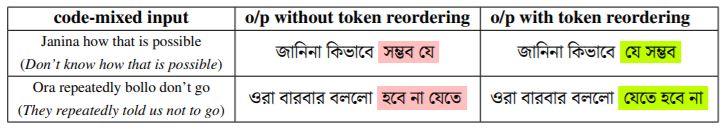}
\caption{Translation with and without token reordering on short snippets.}
\label{fig3}
\end{figure*}

\subsection{Token Reordering}
In several cases, we noticed that the result post joining the outputs from the translation and the transliteration module had grammatical errors, mainly contributed by wrong word ordering other than errors in word forms. To fix the former problem, we created a simple language model based token reordering system. We used the Bengali corpus in TDIL~\footref{TDIL} with 50k sentences to create a trigram and bigram based language model with normalized scores in log space. The system first calculates the normalized log probability of the input sentence. A confusion set, if applicable, is made for each trigram in the sentence. A re-scoring is performed on the sentence by substituting candidates in confusion set. The trigram substitutions (essentially reordering) which results in the best net score is kept. If no alterations are performed by the trigram model, a similar sequence of steps is performed using the bigram model as our final step. This process is inspired from the work in \citet{bryant:2018language}. An example of bigram and trigram ordering is shown in Fig~\ref{fig3}.

\section{Results \& Evaluation}
The scores achieved by our system and Google NMTS (in en-bn setup) is given in Table~\ref{results}. Two variants of our system was tested, one without token reordering (CMT1) and one with (CMT2). Manual scoring (in the range 1-5, low to high quality) of Adequacy and Fluency \cite{banchs:2015adequacy} was done by a bi-lingual linguist, fluent in both En and Bn, with Bn as mother tongue. We can clearly see that our pipeline outperforms GNMT by a fair margin (about 13.34 BLEU, 18.34 TER) and token reordering further improves our system, especially in the case of fluency. Also, for a deeper analysis, we performed two experiments using CMT2. 

\begin{table}[H]
\centering
\begin{tabular}{|c|c|c|c|c|}
\hline
\textbf{Model} & \textbf{BLEU} & \textbf{TER} & \textbf{Adq.} & \textbf{Flu.} \\ \hline
GNMT & 2.44 & 75.09 & 0.90 & 1.12 \\ \hline
CMT1 & 15.09 & 58.04 & 3.18 & 3.57 \\ \hline
CMT2 & 16.47 & 55.45 & 3.19 & 3.97 \\ \hline
\end{tabular}
\caption{Evaluation results.}
\label{results}
\end{table}

\noindent \textbf{\textit{Exp 1.}} We randomly took 100 instances where BLEU score achieved was less than 15. Then we fed this back to our pipeline and collected outputs from each of the modules. We manually associated each of the errors with the respective module causing it, considering the input to it was correct. The results are shown below in Table~\ref{error}. Language tagger being the starting module in our pipeline requires the most improvement for better results followed by the machine translation system and the back-transliteration module. All of these are supervised models and can be improved with more training data.

\begin{table}[H]
\centering
\begin{tabular}{|c|c|}
\hline
\textbf{Module} & \textbf{Contribution} \\ \hline
Language Tagger & 36 \\ \hline
Back Transliteration & 12  \\ \hline
Machine Translation & 25 \\ \hline
\end{tabular}
\caption{Error contribution.}
\label{error}
\end{table}

\noindent \textbf{\textit{Exp 2.}} A linguist proficient in both English and Bengali manually divided our test data into two sets, one where the matrix language was Bengali (M\textsubscript{Bn}) and the other where matrix language was English (M\textsubscript{En}). The size of (M\textsubscript{Bn}) was 1205 and for (M\textsubscript{En}) it was 395. When feeding the sets separately to CMT2, the BLEU and TER score achieved on MBn was 16.98 and 55.02 while on MEn it was 9.3 and 65.11 respectively. This is mainly due to the fact that in our pipeline, the Bn segments are transliterated while En segments are translated and translation has a higher error potential, as compared to transliteration (as shown in Table~\ref{error}). This problem can be easily solved if matrix and embedded languages are identified first, and then passed on to different systems accordingly, i.e. one for (M\textsubscript{Bn}) type, and one for (M\textsubscript{En}) type.

\section{Conclusion \& Future Work}
In this article, we have presented results from our ongoing work on translating code-mixed to monolingual instance. Our system gets a BLEU score of 16.47 on our testing data which is a good starting point. On error analysis, we found out that language identification and translation systems contribute in reduction of BLEU score the most. In the future we would like to add new modules into our pipeline like a matrix-embedded language classifier, an accurate normalization module and replace token reordering with a grammar correction module, something similar to \cite{yuan:2016grammatical}. Our current goals will include improving the language tagger and incorporating context information while translating rather than just segments. Experimenting on chat data which has more noise potential will be interesting as well.

\bibliography{acl2019}
\bibliographystyle{acl_natbib}
\end{document}